\documentclass[10pt,letterpaper]{article}

\usepackage{cogsci}

\cogscifinalcopy %%% Uncomment this line for the final submission

\usepackage{cmap}
\usepackage[T1]{fontenc}
\usepackage[american]{babel}
\usepackage{csquotes}
\usepackage{newtxtext,newtxmath}  %%% TeX Gyre Termes
\usepackage{booktabs}
\usepackage{graphicx}
\usepackage{algpseudocode}
\usepackage{algorithm}

\usepackage{xcolor}
\usepackage{color-edits}
\addauthor[Shane]{shane}{orange}
\addauthor[Ash]{ash}{blue}

\usepackage[
  backend=biber,
  style=apa,
  natbib=true,
  annotation=false,
]{biblatex}
\usepackage{float} %%% Roger Levy added this and changed figure/table placement to [H] for conformity to Word template, though floating tables and figures to top is still generally recommended!

\usepackage[hidelinks]{hyperref}

\ExplSyntaxOn
\NewDocumentCommand{\anonymous}{mm}
  {
    \bool_if:NTF \g_cogsci_final_bool
      { #2 } % final = true
      { #1 } % final = false
  }
\ExplSyntaxOff

\title{When Efficient Communication Explains Convexity}

\author[1 ]{\mbox{Ashvin Ranjan}}
\author[2 ]{\mbox{Shane Steinert-Threlkeld}}
\affil[1]{Paul G. Allen School of Computer Science \& Engineering, University of Washington}
\affil[2]{Department of Linguistics, University of Washington}
\affil[ ]{\{ar31, shanest\}@uw.edu}
\begin{document}

\maketitle

\begin{abstract}

Much recent work has argued that the variation in the languages of the world can be explained from the perspective of efficient communication; in particular, languages can be seen as optimally balancing competing pressures to be simple and to be informative.
Focusing on the expression of meaning---semantic typology---the present paper asks what factors are responsible for successful explanations in terms of efficient communication.
Using the Information Bottleneck (IB) approach to formalizing this trade-off, we first demonstrate and analyze a correlation between optimality in the IB sense and a novel generalization of \emph{convexity} to this setting. In a second experiment, we manipulate various modeling parameters in the IB framework to determine which factors drive the correlation between convexity and optimality.  We find that the convexity of the communicative need distribution plays an especially important role.  These results move beyond showing that efficient communication can explain aspects of semantic typology into explanations for why that is the case by identifying which underlying factors are responsible.

\textbf{Keywords:}
efficient communication; convexity; semantic typology; information bottleneck
\end{abstract}

\section{Introduction}

A central goal in the cognitive science of language centers on explaining semantic typology, the range and limits of variation in how the languages of the world express meaning.  Two prominent questions arise in the pursuit of this goal.  Descriptively: what are the similarities and differences in how languages encode meaning?  Explanatorily: which aspects of this typology are idiosyncratic historical accidents, and which follow from more general cognitive and social factors?

To the descriptive question: it has been argued that word meanings in the world's languages denote \emph{convex} regions in relevant geometric spaces \citep{gardenforsConceptualSpacesGeometry2000, gardenforsGeometryMeaning2014, jagerNaturalColorCategories2010, chemlaConnectingContentLogical2019}.  Convex regions are geometrically well-behaved: for any two points in such a region, any third point between them must also belong to the region.  This creates `smooth' or `natural' borders.  This has been argued to be a kind of `meta-universal', applying to word meanings across many different semantic domains.

To the explanatory question: multiple factors, especially including ease of learning \citep{steinert-threlkeldLearnabilitySemanticUniversals2019, Threlkeld2020, maldonadoLearnabilityConstraintsSemantics2022, douvenLearnabilityNaturalConcepts2025} and efficient communication \citep{kempSemanticTypologyEfficient2018, Zaslavsky2018, zaslavskyLetTalkEfficiently2021, Mollica2021, steinert-threlkeldQuantifiersNaturalLanguage2021, denicIndefinitePronounsOptimize2022, imelEfficientCommunicationAnalysis2026, uegakiInformativenessComplexityTradeOff2023} have been argued to explain why semantic systems are structured the way that they are.  While the efficient communication approach has been successfully applied in many empirical domains, it has largely not been used to explain convexity.  Furthermore, the wide success of the approach calls for a deeper understanding of why it works when it does.

To this end, this paper asks a question connecting the two dimensions: under what conditions can efficient communication explain the convexity of word meanings?  In particular, we approach efficient communication from the perspective of the information bottleneck (IB) framework \citep{Tishby1999, Zaslavsky2018} and report two experiments.

First, we focus on color naming systems, which have been well-studied both from the perspective of convexity \citep{jagerNaturalColorCategories2010, Threlkeld2020} and from the IB framework \citep{Zaslavsky2018}.  First, we introduce a graded measure of \emph{quasi-convexity} which generalizes the degree of convexity used elsewhere \citep{Threlkeld2020, carlssonCulturalEvolutionIterated2024, Koshevoy2025} to the scenario (as in IB) where meanings are probabilistic and show that this measure correlates with IB optimality.  Second, to address our primary question, we introduce smaller, `toy' semantic spaces where we can systematically vary key components of the IB framework, in order to analyze which aspects are essential for producing a correlation between convexity and efficiency.  We find that the relative convexity of the prior distribution over meanings---often called the communicative need distribution---plays an especially pivotal role.

The paper is structured as follows.  In the next section, we introduce both the IB framework and our measure of the degree of convexity.  We then report our results on color naming systems, before turning to the experiment that manipulates IB parameters in a controlled setting.  We conclude by discussing the relation to existing work and ramifications of these results.

\section{Methodology}
\label{sec:methods}

We first present the components of the methodology shared between our two experiments before the results of each of them separately.  Code and data for reproducibility may be found at \anonymous{\url{https://anonymous.4open.science/r/efficiency-convexity-8FCD/}}{\url{https://github.com/CLMBRs/efficiency-convexity}}.

\subsection{Information Bottleneck Encoders}

The Information Bottleneck models communication between a speaker and a listener via efficient compression \citep{Zaslavsky2018}.
The speaker observes a meaning $m\in M$, which is a probability distribution over the given referents in an environment, $U$ (we use $u$ for an item $u\in U$). The speaker then aims to express the meaning $m$ by selecting a word $w$ in their vocabulary $W$. Meanings are conditional probability distributions $p(u|m)$ and speakers are another one, $q(w|m)$, referred to as an encoder $q$.

To calculate the efficiency of a given encoder, we can use two metrics $I_q(W;U)$ and $I_q(M;W)$ \citep{Tishby1999, Zaslavsky2018} which can be thought of as the accuracy and complexity, respectively, of $q$.

\subsubsection{Optimal Encoders}

The accuracy and complexity of an encoder are related to each other: it is impossible to minimize complexity and simultaneously maximize accuracy.  To optimally balance the two measures, the IB framework seeks encoders which minimize
$$
    {\cal F}_\beta[q(w|m)]=I_q(M;W)-\beta I_q(W;U)
$$
where $\beta \geq 0$ is the trade-off parameter.

To calculate the optimal frontier ${\cal F}_\beta[q(w|m)]$ for various values of $\beta$, we used reverse deterministic annealing as described in \citet{Zaslavsky2018}.  Specific implementation details may be found in an Appendix.\anonymous{\footnote{Appendices may be found at the same URL mentioned at the beginning of this \nameref{sec:methods} section.}}{}

\subsubsection{Suboptimal Encoders}

In addition to the optimal encoders (as well as natural language encoders, when available), it is important to sample a wide range of other, suboptimal encoders.
We employ a technique of re-sampling adapted from \citet{Skinner2025} to derive suboptimal encoders while still retaining aspects of the structure of optimal ones. For a given encoder $q$, we create a re-sampled encoder $\hat{q}$ by selecting a random subset of meanings $\hat{M}$ based on the percentage for re-sampling (ranging from 10\% to 100\% in 10\% increments). For each $\hat{m}_i \in \hat{M}$, we randomly draw with replacement $\hat{m}_j$ from $\hat{M}$ and define $\hat{q}(w | \hat{m}_i) = q(w | \hat{m}_j)$. And for all meanings $m_i \notin \hat{M}$, $\hat{q}(w|m_i)=q(w|m_i)$. In addition to controlling for aspects of the structure of the original encoder, we can adjust how different a re-sampled encoder is from the original by adjusting the percentage of meanings re-sampled.

We have avoided using rotation to define suboptimal encoders \citep{Regier2007, Zaslavsky2018, Koshevoy2025}.  Many  of our evaluations depend on the geometry of the CIELab color space; the afore-cited papers rotate along the Hue axis of specified Munsell chips.  This rotation does not commute with the mappings to and from the two color spaces, and so rotating Hue can have deleterious effects on our metrics.

A primary benefit of re-sampling over shuffling or standard rotation (one that is applied within the space that meanings are being evaluated in) is that re-sampling applies a non-invertible function to the set of words $W$ for a given language. If the function applied is instead invertible, as it is with shuffling or standard rotation, the complexity $I(M; W)$ of the encoder would remain constant due to the data-processing inequality.

\subsection{Quasi-Convexity}\label{sec:qc_calc}

Convexity has been generalized from a binary property to a scalar metric (i.e.\ a degree of convexity) for a given subset of points in a space in other work \citep{Threlkeld2020, Koshevoy2025}:
$$
    \text{convexity}(X)=\frac{||X||}{||\text{ConvexHull}(X)||}
$$
where $\text{ConvexHull}(X)$ is the smallest convex set extending $X$.
This definition applies to sets with hard boundaries. However, in the case of IB encoders we have to instead deal with probability distributions, which can be thought of as sets with `soft' boundaries.

We can generalize from here to a notion of quasi-convexity as follows.
Let $p: \Omega \to [0, 1]$ be a p.m.f. or p.d.f. and let $\text{ls}(f, t) := \{ x \mid f(x) \geq t \}$ for $f: X \to \mathbb{R}$, $x \in X$, $t \in \mathbb{R}$ (i.e.\ `ls' for level-set). Then:
\begin{equation}\label{eq:dcon}
\mathsf{dcon}(p) := \int_0^{\max(p(x))} \text{convexity}(\text{ls}(p, t)) dt    
\end{equation}
A finite-sum approximation of $\mathsf{dcon}(p)$ is defined in Algorithm \ref{alg:qc_calc}, which is lightly modified from \citet{Skinner2025}.

\begin{algorithm}
\caption{Quasi-convexity calculation}\label{alg:qc_calc}
\begin{algorithmic}
\Require $steps>0$
\Ensure $0\leq p(x)\ \forall x\in X$, $\sum_{x\in X}p(x)= 1$
\State $mesh \gets \frac{1}{steps}$
\State $level \gets \frac{\max(p(x))}{steps}$
\State $qc \gets 0$
\For{$i\in1,\ldots,steps$}
\State $X_{level} \gets \{x\ |\ p(x)\geq level \times i\}$
\If{$|X_{level}|=0$}
\State $qc \gets qc + mesh$
\Else
\State $X_{hull} \gets ConvHull(X_{level})$ 
\State $qc \gets qc + mesh\times\frac{|X_{level}|}{|X_{hull}|}$
\EndIf
\EndFor
\end{algorithmic}
\end{algorithm}

In order to apply this to a set of conditional probabilities, we take the weighted sum of the convexity of $p(x|y)$ for all $y\in Y$ utilizing Algorithm \ref{alg:qc_calc}. 
Let $p:X\times Y\to[0, 1]$ be a conditional p.m.f. or p.d.f. (e.g.\ an IB encoder). Let $\mathsf{dcon}(p)$ be defined by Equation \ref{eq:dcon}. Then:
\begin{equation}
\label{eq:convexity}
    \mathsf{Convexity}(p)=\sum_{y\in Y}p(y)\mathsf{dcon}(p(\cdot|y))
\end{equation}
In the case of the IB, we will focus primarily on the $\text{dcon}(q(m | w))$ and the weighted average across $q(w)$ in (\ref{eq:convexity}).

\section{Experiment 1: Color Naming Systems}\label{sec:exp1}

\subsubsection{Setup}\label{sec:exp1_setup}

Our first experiment focuses on color naming systems, using the IB environment detailed in \citet{Zaslavsky2018}. This environment utilizes data from the World Color Survey (WCS), which displayed 330 Munsell color chips to participants in order to gather natural language data for color naming \citep{berlinBasicColorTerms1969, Cook2005}. These chips are the referents $U$ and meanings are Gaussian distributions over the color chips in CIELab color space\footnote{The conversion from the WCS color space to CIELab utilizes the data from \href{https://linguistics.berkeley.edu/wcs/data.html}{linguistics.berkeley.edu/wcs/data.html}}, with a standard deviation of 64. Furthermore, we use the least-informative prior over meanings described in \citet{Zaslavsky2018}.

\subsubsection{Procedure}

We calculated the optimal encoders using an implementation of IB and the same 1501 values of $\beta$ and reverse annealing described in \citet{Zaslavsky2018}.\footnote{The values of $\beta$, along with additional data which we have used for this experiment, are available at \href{https://github.com/nogazs/ib-color-naming}{github.com/nogazs/ib-color-naming}.}
Natural language data from WCS is turned into encoders by using frequency information. Specific details can be found in an Appendix. 
We re-sample the optimal encoders to generate suboptimal encoders sampling at percentages ranging from 10\% to 100\%, with 10\% increments. 

Because convexity is a property of word meanings, we measure quasi-convexity by applying Equation~\ref{eq:convexity} to the distribution $q(m|w)$, which is defined by Bayes' rule \citep{Zaslavsky2018}.  We measure the optimality of an encoder as the negative Epsilon measure from \citet{Zaslavsky2018}.

\subsubsection{Results}

\begin{figure}[t]
  \includegraphics[width=\columnwidth]{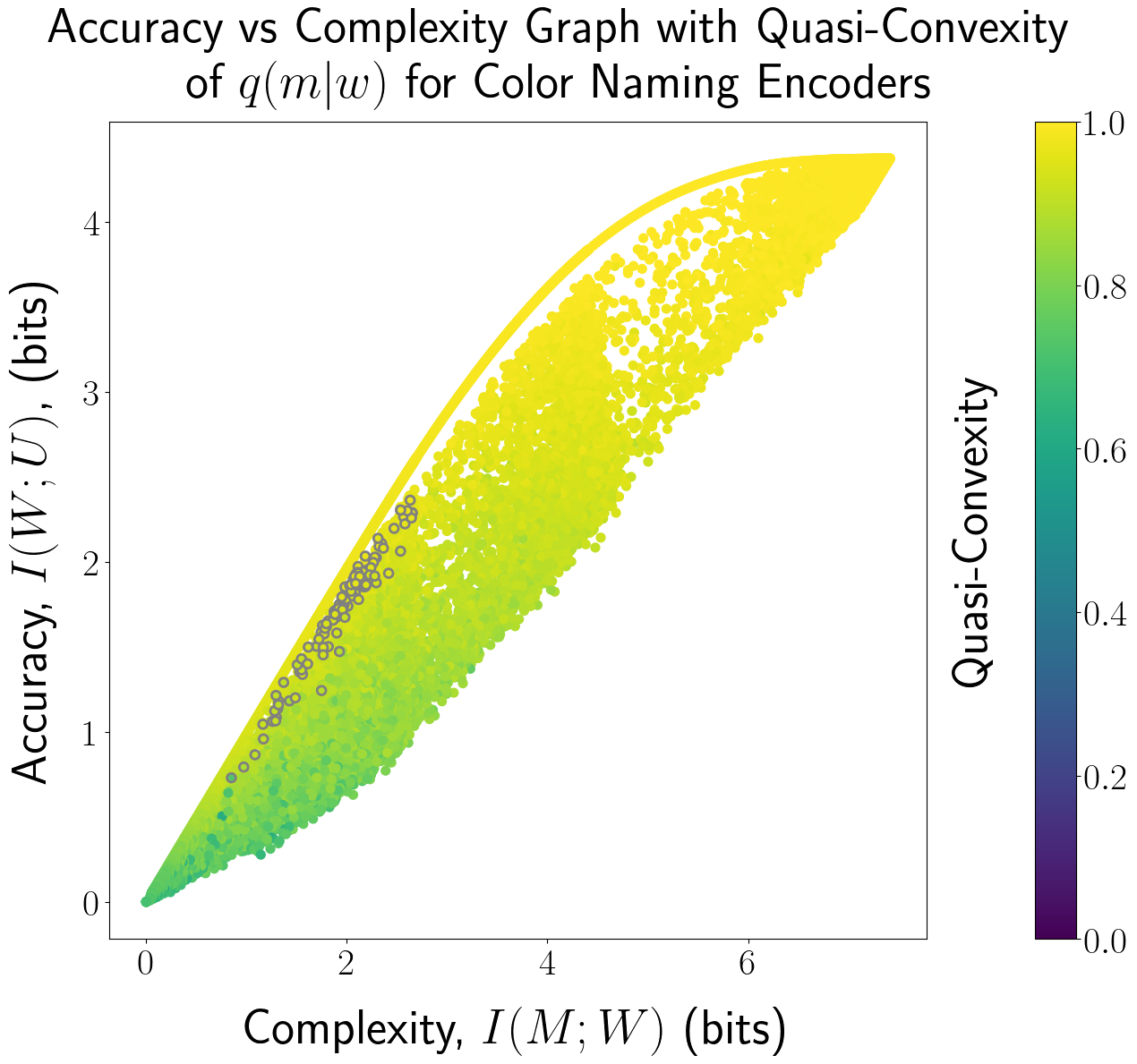}
  \caption{Accuracy vs complexity trade-off for color naming. Color represents quasi-convexity, ranging from purple (0.0 convexity) to yellow (1.0 convexity). Points with a gray outline represent a natural language.}
  \label{fig:graphs_color}
\end{figure}

Figure~\ref{fig:graphs_color} visualizes the quasi-convexity of both distributions for optimal, suboptimal, and natural encoders. 
Pearson correlations with optimality, complexity, and accuracy are reported in Table~\ref{table:color_coeffs}.

\begin{table}[ht]
\centering
\begin{tabular}{cccc}
\toprule
Distribution & Optimality & $I(M; W)$ & $I(W; U)$ \\
\midrule
$q(m|w)$ & 0.221 & 0.797 & 0.885\\
\bottomrule
\end{tabular}
  \caption{Pearson's correlation coefficients between the quasi-convexity of $q(m|w)$ and Optimality, Complexity ($I(M; W)$), and Accuracy ($I(W;U)$) in color naming. The $p$ values for all correlation coefficients are less than $0.00$.
  }
  \label{table:color_coeffs}
\end{table}

We observe a small but non-trivial positive correlation between optimality and quasi-convexity, suggesting that as languages become more optimal, they also become more convex. We observe much larger correlations, however, between quasi-convexity and the complexity and accuracy metrics.

In order to explore the relationships between quasi-convexity and all of these metrics in more detail, we fit a mixed-effects regression model predicting convexity as a function of encoder type (natural, optimal, suboptimal) and ``base type'' (natural or optimal; describing which type of encoder a suboptimal one is derived from), with a random effect for the base encoder. A likelihood ratio test with a standard linear regression omitting the random effect showed that including the random effect significantly improves model fit ($\chi^2(1) = 16911.75, p \approx 0$).

Across 17,721 observations from 1,611 item groups (all groups sized 11, for base encoder and 10 variants of it), both predictors showed significant effects. Compared to natural language encoders (the baseline for the encoder type variable), optimal encoders showed slightly lower convexity scores ($\beta = -0.032, \text{SE} = 0.004, z = -7.49, p < .001$)---likely due to the lower-convexity optimal encoders in the low-accuracy and low-complexity region of the plane---and suboptimal items showed substantially lower scores ($\beta = -0.106, \text{SE} = 0.004, z = -25.30, p < .001$). Items derived from optimal base encoders showed significantly higher convexity values than those derived from natural base encoders ($\beta = 0.066, \text{SE} = 0.007, z = 9.91, p < .001$). The random intercept exhibited a small but non-zero variance component (0.004), indicating meaningful between-item variability.  These results show that suboptimal encoders do in fact have lower convexity scores, but that those derived from optimal encoders are slightly more convex than those derived from natural language encoders.

Finally, a multiple linear regression was conducted to examine how optimality, complexity, and accuracy jointly predict convexity.  This analysis focused on $q(m|w)$ only, since it has the higher correlations with these factors. The model included all two-way and three-way interactions among these continuous predictors, as well as fixed effects for item type and base type. The overall model explained a large amount of variance in convexity scores ($R^2 = .93$, $F(10; 17{,}710) = 23{,}890$, $p < .001$).

All three continuous predictors showed strong main effects. Convexity increased with greater optimality ($\beta = 0.225$, $SE = 0.003$, $t = 80.72$, $p < .001$), complexity ($\beta = 0.16$, $SE = 0.002$, $t = 87.36$, $p < .001$), and (to a smaller degree) accuracy ($\beta = 0.014$, $SE = 0.001$, $t = 11.30$, $p < .001$). These effects, however, were moderated by their interactions. Small negative interactions between optimality and both complexity ($\beta = -0.017$, $p < .001$) and accuracy ($\beta = -0.006$, $p < .001$) show that the positive effect of optimality gets attenuated at higher levels of the latter two factors.  The effect of complexity similarly decreased as accuracy increased ($\beta = -0.030$, $p < .001$). The three-way interaction among optimality, complexity, and accuracy ($\beta = 0.003$, $p = 0.092$) partially offsets these negative interactions but only approaches significance.

By contrast, categorical predictors exhibited minor effects. Neither the optimal type nor the base type had significant effects once the continuous predictors and their interactions were included, although suboptimal items showed slightly lower convexity scores than the natural languages ($\beta = -0.006$, $p = .015$).  Similarly, encoders derived from optimal languages were slightly less convex than ones derived from natural languages ($\beta = -0.002$, $p = 0.026$). Overall, the results indicate variance in convexity can be explained by a complex interaction between optimality, complexity, and (to a lesser extent) accuracy, but natural languages are more convex than suboptimal ones even when accounting for those factors.

\begin{figure*}[ht]
  \includegraphics[width=\textwidth]{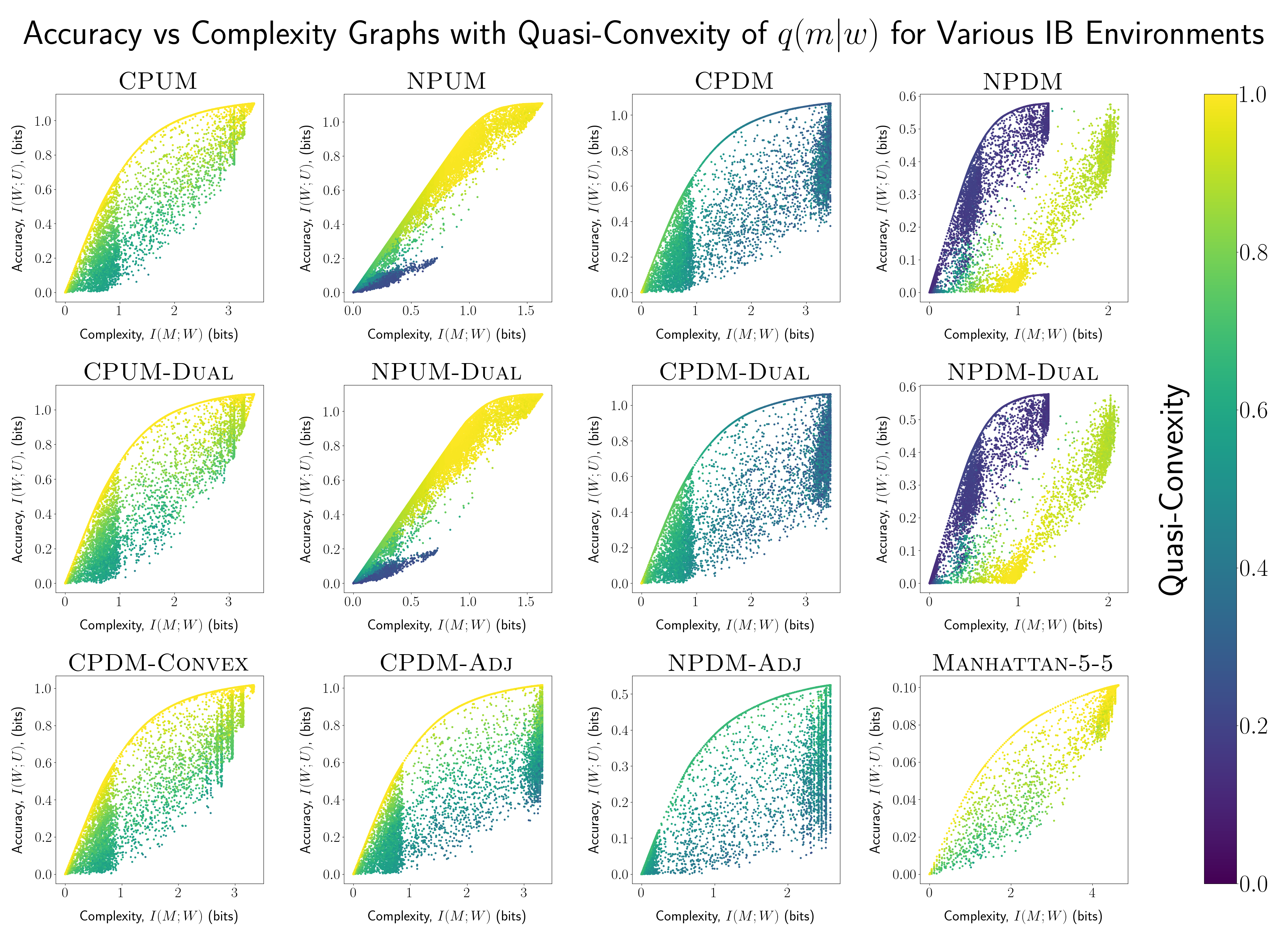}
  \caption{Accuracy vs complexity trade-off graphs for all environments. Color corresponds to quasi-convexity of $q(m|w)$.}
  \label{fig:graphs_qmw}
\end{figure*}

\section{Experiment 2: Manipulating IB}\label{sec:exp2}

Having demonstrated and analyzed a correlation between convexity and optimality in color naming, 
the second experiment aims to understand the factors that explain this correlation by manipulating various components of the IB framework in a simplified setting, creating various environments.

\subsubsection{Base Environments}

We begin with four base environments, focusing on convexity of the priors and uniqueness of meanings.  These all share the set of referents $U = \{0, \dots , 10\}$ and meaning distributions:
$$p(u|m)\propto\phi\left(u,|m|,\frac{9}{4}\right)$$
where $\phi(u,\mu,\sigma^2)$ is the p.d.f.\ of the normal distribution.

The four environments, with explicit definitions of the meaning space and prior given in Table~\ref{table:base_env}, are: CPUM (convex priors, unique meanings), NPUM (non-convex priors, unique meanings), CPDM (convex prior, duplicate meanings), and NPDM (non-convex prior, duplicate meanings).  Convex prior environments have uniform priors, while non-convex ones are skewed heavily towards the `edge' of the environment.  DM environments have two identical meanings (thanks to the $|m|$ in the definition of $p(u|m)$) for each referent $u$.

\begin{table}[ht]
    \centering
    \begin{tabular}{ccc}
    \toprule
         Environment & $M$ & $p(m)$ \\
         \midrule
         \textsc{CPUM} & $\{0,\ldots,10\}$ & $\frac{1}{||M||}$ \\
         \textsc{NPUM} & $\{0,\ldots,10\}$ & $
    \begin{cases}
    0.455 & m = 0, \\
    0.455 & m = 10, \\
    0.01 & \text{otherwise}
    \end{cases}$\\
         \textsc{CPDM} & $\{-10,\ldots,10\}$ & $\frac{1}{||M||}$ \\
         \textsc{NPDM} & $\{-10,\ldots,10\}$ & $
    \begin{cases}
    0.455 & m = -10, \\
    0.455 & m = 10, \\
    0.01 & \text{otherwise}
    \end{cases}$ \\
    \bottomrule
    \end{tabular}
    \caption{The meanings $M$ and prior $p(m)$ for the base environments. $U$ and $p(u|m)$ are shared across these environments.}
    \label{table:base_env}
\end{table}

We additionally explore the following additional environments, to more thoroughly explore the IB parameter space.

\subsubsection{\textsc{-Dual} Variations}  For each base environment, we introduce a \textsc{-Dual} variant, with $U=M=\{-10,\ldots,10\}$ and
$$p(u|m)\propto\left(\phi\left(u,m,\frac{3}{2}\right) + \phi\left(u,-m,\frac{3}{2}\right)\right)/2$$
This distribution is a mixture of two normal distributions with peaks at $u$ and $-u$, which means that $p(u|m)=p(-u|m)$.
\

\subsubsection{\textsc{CPDM-Convex}} This resembles \textsc{CPDM}, but the priors are heavily weighted to negative values in $M$. This means that the priors are still convex, but are not uniform. Specifically:
\[
p(m)=\begin{cases}
    0.099 & m < 0, \\
    \frac{0.01}{11} & \text{otherwise}
\end{cases}
\]

\subsubsection{\textsc{CPDM-Adj}} This resembles \textsc{CPDM}, but with duplicate meanings adjacent to each other in the space. Specifically: $M=\{0,\ldots,19\}$ and $p(u|m)\propto \phi\left(u,\left\lfloor\frac{m}{2}\right\rfloor,\frac{3}{2}\right)$

\paragraph{\textsc{NPDM-Adj}} Like \textsc{CPDM-Adj}, but groups of identical meanings are non-convex.  Specifically:
$M=\{0,\ldots,17\}$, $p(u|m)\propto \phi\left(u,\left\lfloor\frac{m}{3}\right\rfloor,\frac{3}{2}\right)$, and
$$
p(m) = \begin{cases}
        \frac{0.01}{6} & m -1 \equiv 0\mod 3 \\ 
        \frac{0.495}{6} & \text{otherwise}
    \end{cases}
$$
Duplicate meanings come in contiguous blocks of three (thanks to $\lfloor m/3 \rfloor$), but the middle one has substantially lower prior probability than the outer two.

\paragraph{\textsc{Manhattan-5-5}} The referents form a $5\times5$ grid, with $p(u|m)$ being the normalized Manhattan distance from $m$ to $u$. Specifically: $U = M = \{0, \dots, 4\} \times \{0, \dots , 4 \}$, the prior over meanings is uniform, and $p(u|m) \propto \text{dist}(u, m) + 1$, where \text{dist} is the Manhattan distance and 1 is added to avoid 0.

\subsection{Procedures}\label{sec:exp2_procedures}

For each environment, we followed the same reverse deterministic annealing procedure as in the previous experiment to estimate the optimal frontier, and used the same re-sampling procedure and percentages to generate suboptimal encoders.  Note that there are no natural language encoders in this experiment, since we are using hypothetical environments to test the correlation between convexity and optimality.

\subsection{Results}\label{sec:exp2_results}

Figure~\ref{fig:graphs_qmw} visualizes quasi-convexity of $q(m|w)$ for optimal and suboptimal encoders in each environment. 
Pearson correlations with optimality, complexity, and accuracy are reported in Table~\ref{table:coeffs_qmw}.  Several trends can be observed from this data.

\begin{table}[ht]
\centering
\begin{tabular}{lccc}
\toprule
\multicolumn{1}{c}{Environment} & Optimality & $I(M; W)$ & $I(W; U)$ \\ 
\midrule
\textsc{CPUM} & 0.860 & -0.169 & -0.074\\
\textsc{NPUM} & 0.238 & 0.791 & 0.9\\
\textsc{CPDM} & 0.751 & -0.907 & -0.879\\
\textsc{NPDM} & -0.935 & 0.736 & 0.295\\ \midrule
\textsc{CPUM-Dual} & 0.855 & -0.193 & -0.092\\
\textsc{NPUM-Dual} & 0.226 & 0.791 & 0.901\\
\textsc{CPDM-Dual} & 0.752 & -0.907 & -0.872\\
\textsc{NPDM-Dual} & -0.934 & 0.736 & 0.289\\  \midrule
\textsc{CPDM-Convex} & 0.869 & -0.381 & -0.257\\
\textsc{CPDM-Adj} & 0.882 & -0.619 & -0.423\\
\textsc{NPDM-Adj} & 0.889 & -0.641 & -0.380 \\
\textsc{Manhattan-5-5} & 0.811 & -0.242 & -0.188\\
\bottomrule
\end{tabular}
  \caption{
  Pearson's correlation coefficients between the quasi-convexity of $q(m|w)$ and Optimality, Complexity ($I(M; W)$), and Accuracy ($I(W;U)$) for environments in Experiment 2. The $p$ values for all correlation coefficients are less than $0.00$.}
  \label{table:coeffs_qmw}
\end{table}

First, convex priors play a crucial role: every environment with convex priors (those starting with \textsc{CP-}) exhibits a strong correlation between optimality and quasi-convexity.  These environments have nearly perfectly convex encoders.  Unlike the color naming case, these environments also exhibit \emph{negative} correlations between quasi-convexity and the IB accuracy and complexity metrics. \textsc{Manhattan-5-5}, which is roughly a two-dimensional analog of \textsc{CPUM}, exhibits the same pattern.

Crucially, however, the optimal languages in the duplicate meaning (-DM) environments are not, in general, fully convex. The CPDM environment, in particular, has convex priors but non-convex optimal meanings. This echoes concurrent results from \citet{imel2026convexityefficiencysemanticsystems}, whose construction of IB-optimal but non-convex languages resembles the CPDM construction but with a circular instead of a linear symmetry.  Although the optimal encoders in this setting are not convex, and they become less convex as one moves along the optimal frontier (also evidenced by strong negative correlations with complexity and accuracy), there is still a correlation between optimality and convexity.

Furthermore, each of the \textsc{-Dual} environments show similar behavior---both visually and in terms of the three correlations---as their base counterpart.  This suggests that adding dual meanings to the environment does not dramatically change the behavior of IB, consistent with recent results showing that IB complexity is not sensitive to synonymy \citep{bruneau-bongardAssessingPressuresShaping2025}. Along with \textsc{Manhattan-5-5}, the behavior of the \textsc{-Dual} environments indicates that the convexity of the $p(u|m)$ does not play a dramatic role in the quasi-convexity of $q(m|w)$  for the environment's encoders.

The non-convex prior environments exhibit an important distinction: all \textsc{NPUM} environments exhibit a positive but weaker correlation between optimality and quasi-convexity and, unlike the \textsc{CP-} variants, positive correlations with the two IB metrics.  \textsc{NPDM} and its dual variant show a strong negative correlation between optimality and convexity, reflected in the optimal encoders there all having nearly-0.0 quasi-convexity.

These results connect to the color naming scenario: the correlation coefficients for the \textsc{NPUM} environments are quite similar to the coefficients for the $q(m|w)$ distributions in the color naming environment in Table~\ref{table:color_coeffs}. This is further substantiated by the fact that all meanings are unique in the color naming environment and that $\mathsf{dcon}(p(m))=0.705$, which, while more convex than the prior distribution of \textsc{NPUM} (where $\mathsf{dcon}(p(m))=0.206$), is still relatively non-convex.

\section{Discussion}

Through two experiments connecting the information bottleneck framework for efficient communication to a generalization of the convexity universal, we have shown the following.  In color naming, there is a non-trivial correlation between convexity and optimality, suggesting that efficient communication can partially explain convexity.  We also found that natural languages are slightly more convex than other suboptimal encoders, even when controlling for optimality, accuracy, and complexity. By manipulating various components of the IB framework, we found that the convexity of the prior distribution (with unique meanings) is the most significant factor in driving a correlation between convexity and optimality.

Concurrently with this work, \citet{Koshevoy2025} have also explored the connection between convexity and optimality in color naming, finding a strong and convincing correlation.  Our work, while consistent with their results, differs in a few important ways.  They rely on the partition-generating algorithm and measure of degree of convexity from \citet{Threlkeld2020}, both of which apply to meanings with `hard' boundaries.  One of our main contributions generalizes the notion of the degree of convexity to probabilistic encoders of the kind used in IB.  Similarly, their primary analysis operationalizes complexity and accuracy in slightly different ways, while we focus only on the IB framework.  To this end, \citet{bruneau-bongardAssessingPressuresShaping2025} compare multiple operationalizations of complexity (including the earlier degree of convexity) as well as combining more factors than just complexity and accuracy, in the color naming domain. Also currently with this work, \citet{imel2026convexityefficiencysemanticsystems} analyze the relationship between convexity and optimality using the same notion of convexity as the two works above. Like the present paper, they utilize the IB system for analyzing encoders; they show, among other things, that neither IB-optimality nor convexity entail the other and argue that the former is more fundamental.  In addition to the more fitting definition of degree of convexity, the present paper also moves beyond only color naming in using constructed domains to tackle the question of which factors underlie the observed correlation between convexity and optimality.

\citet{carlssonCulturalEvolutionIterated2024} show that a model of cultural evolution capturing both iterated learning and communication amongst neural agents produces color naming systems that are both IB-optimal and human-like.  Implicit in their studies is also the production of systems which are convex but inefficient, suggesting that convexity is necessary but not sufficient for distinguishing natural from unnatural color naming systems.   Our work---including cases with dual meanings where convexity and efficiency are anti-correlated---complements this perspective well and extends it beyond color naming in trying to identify \emph{when} it is that efficient systems are also convex.

Future work can expand the results here along several axes.  The analyses in Experiment~2 can be made more systematic by manipulating factors like convexity of the priors in a continuous manner and attempting to predict the strength of the correlation between convexity and optimality from them.  Similarly, these analyses can be applied to a wider range of semantic spaces, including others that have been analyzed in the IB framework.  Finally, it may be possible to prove analytic results in the vicinity of our experimental results. Our CPDM case---as well as the similar construction in \citet{imel2026convexityefficiencysemanticsystems}---shows that convex priors does not suffice.  One candidate, refined conjecture: with convex priors and \emph{unique} meanings, all IB-optimal encoders are convex.  More generally, the methods here could be used to help adjudicate between different operationalizations of complexity and accuracy in efficient communication: if some more robustly generate a correlation with semantic universals across other semantic domains and manipulations of other parameters, it suggests that they are more likely to be explanatory factors.

By analyzing which factors underlie a connection between efficient communication and a prominent semantic universal, this paper takes a first step in deepening the kinds of explanations in computational approaches to semantic typology.

\printbibliography

\clearpage 

\appendix
\section{Additional Notes About Reverse Deterministic Annealing}\label{sec:app_a}

Reverse deterministic annealing is a method which starts with a large value of $\beta$ and applies the information bottleneck method from \citet{Tishby1999} until convergence to a fully accurate encoder. That being, an encoder where each meaning has one unique word which corresponds to it. The resulting output is then used as the starting encoder for a lower value of $\beta$. This repeats until the final, lowest value of $\beta$ \citep{Zaslavsky2018}. An additional note is that if, for a given $w\in W$, $q(w|m)= 0$ for all $m\in M$, $w$ is removed from the encoder \citep{Piasini2021}.

In our implementation of reverse deterministic annealing, the range of $\beta$ values used is the same as in \citet{Zaslavsky2018}, which goes from 0 to $2^{13}$. Furthermore, it is important to note that the use of the algorithm defined in \citet{Tishby1999} adds an additional consideration is introduced which requires $p(w)$, $p(m)$, and $p(u|m)$ to be full support. This is because the Kullback–Leibler divergence between $p(u|m)$ and $q(u|w)$ is taken multiple times, and all probabilities must be non-zero in calculating the Kullback–Leibler divergence.

\section{Further Details on Rotation for Supoptimal Encoders}

Previous works involving color naming have utilized the technique of rotation in order to create variations on encoders either optimal or from natural languages \citep{Zaslavsky2018,Regier2007,Koshevoy2025}.

\begin{figure}[ht]
  \includegraphics[width=\columnwidth]{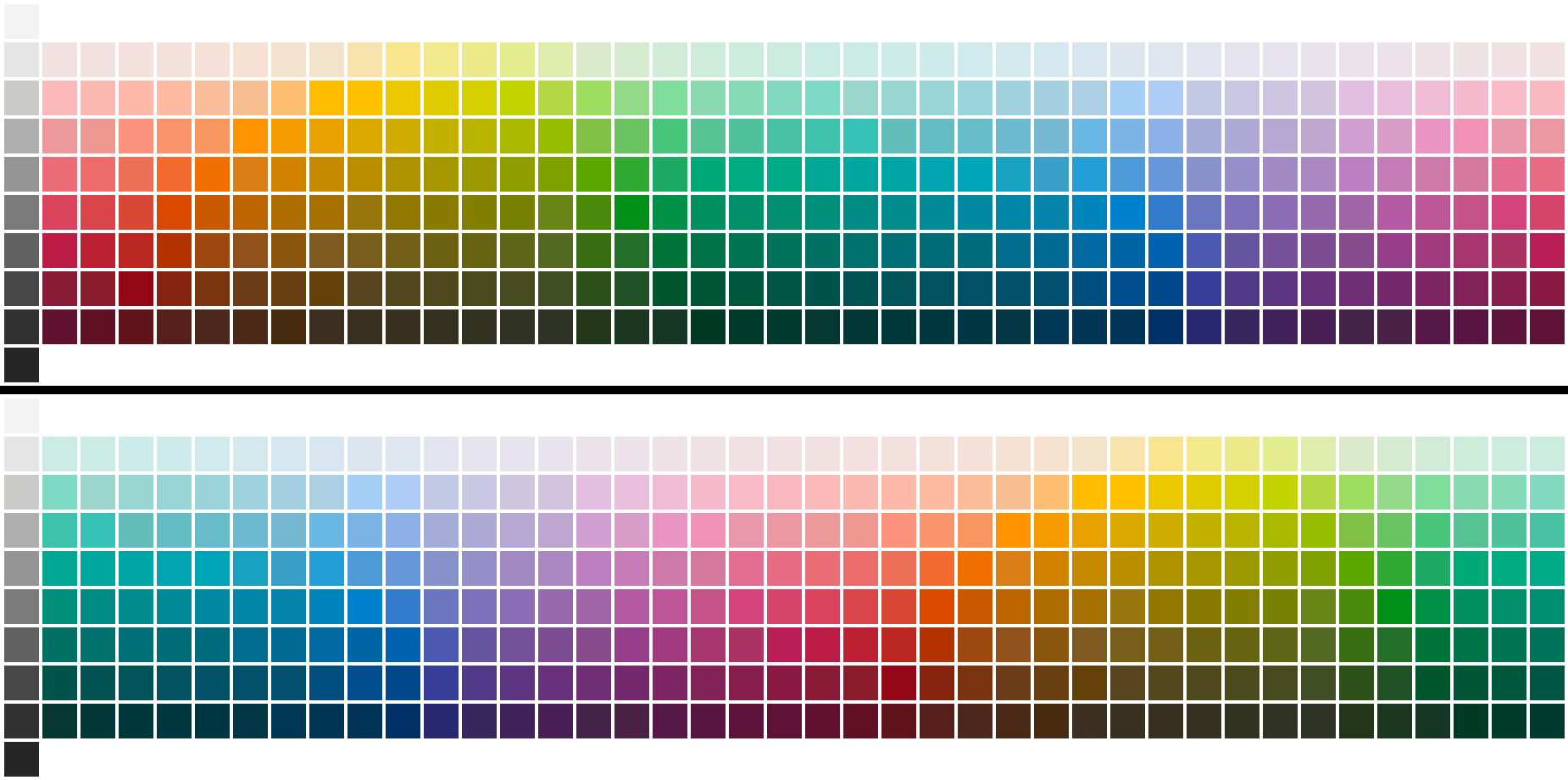}
  \caption{The stimulus grid for the World Color Survey and its rotation by 20. The top image is the stimulus grid used by the World Color Survey, with the vertical axis being the Value and the horizontal axis being the Hue. The bottom image is the same but all chips with a non-zero Hue value are rotated along the Hue axis by 20.}
  \label{fig:colors_munsell}
\end{figure}

This technique originates from \citet{Regier2007}, where it was used to create suboptimal languages which could then be evaluated. However that paper, and others following it \citep{Zaslavsky2018,Koshevoy2025}, all rotate by only changing the Hue value of the given chip. An example of this is shown in Figure \ref{fig:colors_munsell}. 

\begin{figure}[ht]
  \includegraphics[width=\columnwidth]{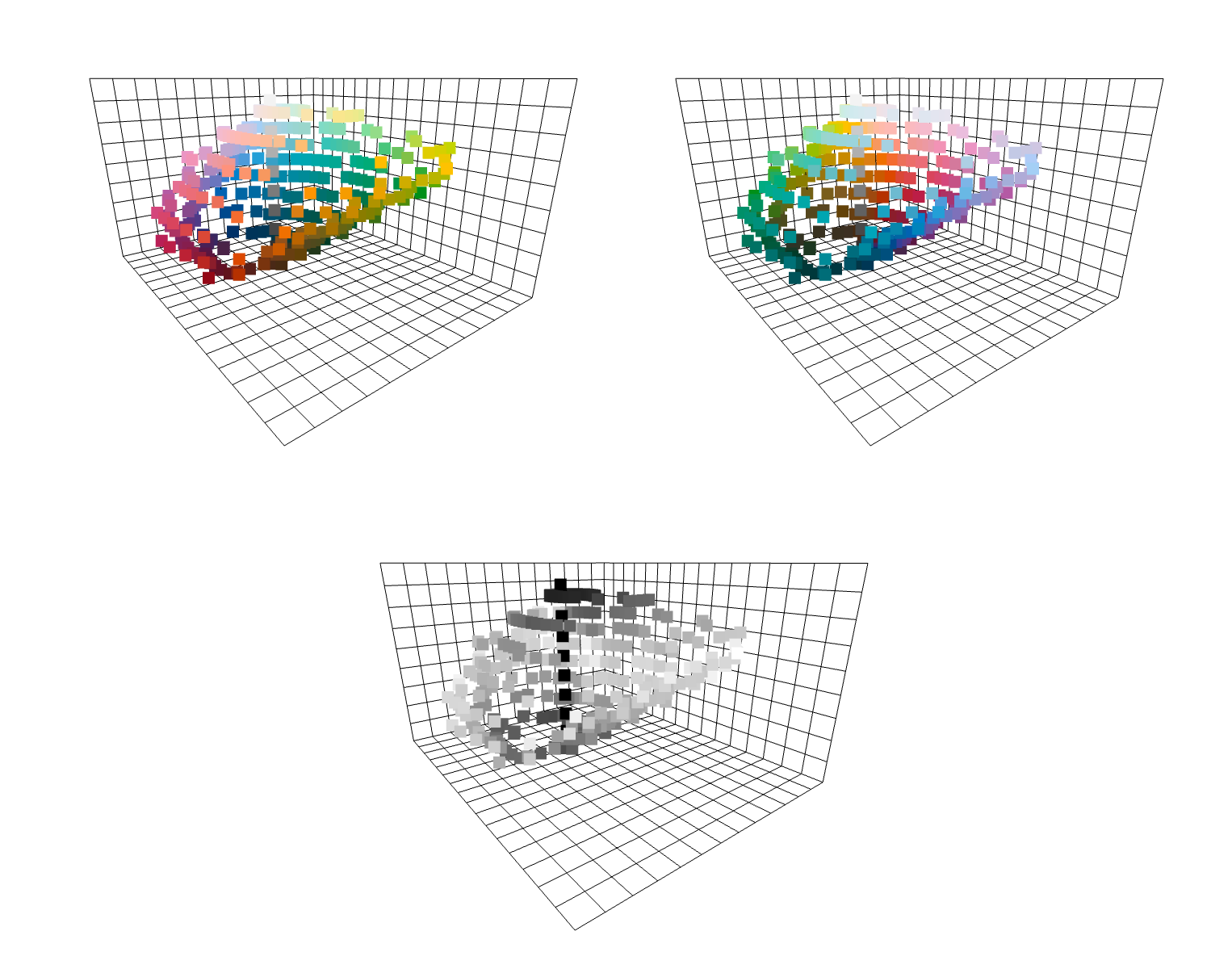}
  \caption{Visualizations of the colors in CIELab space. The top left image shows the normal distribution of colors in CIELab space, with each point being colored corresponding to its location. The top right image shows the same points however colored in correspondence to their color after a rotation of 20 in the Hue axis of the WCS coordinates (see Figure \ref{fig:colors_munsell}). The bottom figure shows the location difference between the colors for each point in the top figures, with black corresponding to no change and white corresponding to maximal change. Note that the planes shown are not the L*a*, a*b*, and L*b* planes, but instead parallel to help with viewing.}
  \label{fig:colors_cielab}
\end{figure}

These chips' current placement is based on the Munsell system \citep{Cook2005}, however evaluation of these systems are usually done with the chips' CIELab coordinates \citep{Regier2007,Zaslavsky2018,Koshevoy2025}. This causes the rotation to have a non-trivial effect on the CIELab location of the chip as can be seen in Figure \ref{fig:colors_cielab}.

Because of this, we have decided to not utilize rotation to generate suboptimal encoders within this paper, as the transformation applied in the color domain via rotation done in previous work is not easily representable in other domains.

\section{Deriving Encoders for Natural Languages from World Color Survey Data}\label{sec:app_b}

The WCS data provides a table which contains term responses questioned speakers responded with for each chip for each language the survey covered  \citep{Cook2005}. We are able to use this table to create a $q(w|m)$ distribution. For a given language, each speaker will have their own response to a given chip $m$ (note that their response $w$ is based on the meaning they perceive, $m$, for the given chip). As such:
$$q(w|m)=\frac{\#w\text{ responses for chip }m}{\#\text{speakers for language}}$$

\section{Full Regression Results for Experiment 1}
\label{app:regressions}

\begin{table*}[ht]
\centering
\caption{OLS regression predicting convexity}
\label{tab:convexity-regression}
\begin{tabular}{lcccc}
\toprule
Predictor & $\beta$ & $SE$ & $t$ & $p$ \\
\midrule
Intercept & 0.7420 & 0.002 & 328.058 & $< .001$ \\

Type (optimal) & 0.0035 & 0.002 & 1.514 & 0.130 \\
Type (suboptimal) & $-0.0055$ & 0.002 & $-2.423$ & 0.015 \\
Base type (optimal) & $-0.0016$ & 0.001 & $-2.226$ & 0.026 \\

Optimality & 0.2225 & 0.003 & 80.715 & $< .001$ \\
Complexity & 0.1602 & 0.002 & 87.362 & $< .001$ \\
Accuracy & 0.0144 & 0.001 & 11.298 & $< .001$ \\

Optimality $\times$ Complexity & $-0.0174$ & 0.001 & $-19.058$ & $< .001$ \\
Optimality $\times$ Accuracy & $-0.0063$ & 0.001 & $-4.661$ & $< .001$ \\
Complexity $\times$ Accuracy & $-0.0304$ & $<0.001$ & $-112.406$ & $< .001$ \\

Optimality $\times$ Complexity $\times$ Accuracy & $-0.0003$ & $<0.001$ & $-1.687$ & $0.092$ \\
\midrule
Observations & \multicolumn{4}{c}{17{,}721} \\
$R^2$ & \multicolumn{4}{c}{0.931} \\
Adjusted $R^2$ & \multicolumn{4}{c}{0.931} \\
\bottomrule
\end{tabular}

\begin{flushleft}
\footnotesize
\textit{Note}. Coefficients are unstandardized. Type and base type are dummy-coded with the natural condition as the reference level. All continuous predictors were entered with full two-way and three-way interactions. $p$ values are two-tailed.
\end{flushleft}
\label{table:reg_results}
\end{table*}

The full regression results discussed in the results for Experiment 1 are detailed in Table \ref{table:reg_results}.

\end{document}